\documentclass{article}

\usepackage{microtype}
\usepackage{graphicx}
\usepackage{booktabs}
\usepackage{tabularx}
\usepackage{array}
\usepackage{fancyvrb}
\usepackage{fvextra}
\usepackage{hyperref}
\usepackage{amsmath}
\usepackage{amssymb}
\usepackage{mathtools}
\usepackage{amsthm}
\usepackage[capitalize,noabbrev]{cleveref}

\usepackage[accepted]{icml2025}

\theoremstyle{plain}

\theoremstyle{definition}

\theoremstyle{remark}

\graphicspath{{figures/}}

\icmltitlerunning{LLM Framework for Discovering Major Mathematical Conjectures}

\begin{document}

\twocolumn[
\icmltitle{LLM Framework for Discovering Major Mathematical Conjectures: AI's Quest for the Next Riemann Hypothesis}

\begin{icmlauthorlist}
\icmlauthor{Alizer Wong}{pku}
\icmlauthor{Zixin Zeng}{gdut}
\icmlauthor{Yi Tan}{gdutie}
\icmlauthor{Wenyuan Li}{hkou}
\icmlauthor{Xuhang Chen}{hzu}
\icmlauthor{Xingru Lai}{scnu}
\icmlauthor{Yang Shi}{gdut}
\icmlauthor{Liangsi Lu}{gdut1}
\icmlauthor{Yanhui Chen}{gdut}
\end{icmlauthorlist}

\icmlaffiliation{pku}{School of Computer Science, Peking University}
\icmlaffiliation{gdut}{School of Computer Science and Technology, Guangdong University of Technology}
\icmlaffiliation{gdutie}{School of Information Engineering, Guangdong University of Technology}
\icmlaffiliation{hkou}{Hokkaido University}
\icmlaffiliation{hzu}{School of Computer Science and Engineering, Huizhou University}
\icmlaffiliation{scnu}{School of Artificial Intelligence, South China Normal University}
\icmlaffiliation{gdut}{School of Computer Science and Technology, Guangdong University of Technology}
\icmlaffiliation{gdut1}{Guangdong University of Technology}
\icmlaffiliation{gdut}{School of Computer Science and Technology, Guangdong University of Technology}
\icmlcorrespondingauthor{Alizer Wong}{aliiiiezer@gmail.com}
\icmlcorrespondingauthor{Zixin Zeng}{zengzixin@mails.gdut.edu.cn}
\icmlcorrespondingauthor{Yi Tan}{tyyeahhhhh@outlook.com}
\icmlcorrespondingauthor{Wenyuan Li}{wenyuan@lmd.ist.hokudai.ac.jp}
\icmlcorrespondingauthor{Xuhang Chen}{xuhangc@hzu.edu.cn}
\icmlcorrespondingauthor{Xingru Lai}{20254002143@m.scnu.edu.cn}
\icmlcorrespondingauthor{Yang Shi}{sudo.shiyang@gmail.com}
\icmlcorrespondingauthor{Liangsi Lu}{lu.liangsi.cn@gmail.com}
\icmlcorrespondingauthor{Yanhui Chen}{chenyanhui91@mails.gdut.edu.cn}

\icmlkeywords{mathematical conjecture discovery, large language models, Lean, Mathlib, semantic validation, formal verification}

\vskip 0.3in
]

\printAffiliationsAndNotice{}

\begin{abstract}
Major mathematical conjectures still depend heavily on expert intuition, so a unified method for the systematic generation and validation of conjectures with substantial mathematical potential remains unavailable. We present a three stage pipeline for major conjecture discovery, with region search from explicit local evidence modules, reflective validation for foundationality, novelty, and potential significance, and formal validation in Lean 4 and Mathlib. The objective is the discovery of mathematical problems with high problem taste, namely problems whose proofs could reorganize the language of a research area and provide durable help to human mathematical research. Experiments on twenty candidates show stable passage from natural language to formal checks, with twenty out of twenty candidates passing Lean parsing and type checking, twenty out of twenty candidates not directly absorbed by exact?, twenty out of twenty candidates not automatically discharged by aesop, and no explicit duplicates or near duplicates.

\end{abstract}

\section{Introduction}
\label{sec_intro}

Major mathematical conjectures do not merely form a collection of unresolved problems. Such conjectures organize mathematical knowledge and guide the allocation of long term research effort. The continuing centrality of the Riemann hypothesis, Goldbach type statements, and the conjecture of Birch and Swinnerton Dyer is not explained only by proof difficulty. Their importance is tied to compressed structural tensions that connect discrete phenomena with analytic objects, local constraints with global solvability, and previously separate mathematical languages. Representative discussions appear in the official statement of the Clay Mathematics Institute \citep{clay2025riemann} and in the classical work of \citet{hardy1923expression}. From this perspective, the formulation of major conjectures is itself a methodological problem. If major conjectures shape the organization of mathematical knowledge, then the systematic generation and validation of conjectures with major potential should not be treated as an irreducibly informal act of expert intuition. The target is not any novel question. The target is a question with high problem taste and with the potential to help human mathematical research in a lasting way after a proof becomes available.

The historical formation of major conjectures follows a recognizable pattern. Empirical regularities often appear before structural explanations. High impact statements are frequently attached to high fidelity proxy objects. Weak versions, boundary results, and main term formulae often mature before a terminal assertion is stated. The account of plausible reasoning from \citet{polya1954mathematics} and the proof and refutation cycle analyzed by \citet{lakatos1976proofs} both support this view. The central difficulty in major conjecture discovery is therefore not the production of a new sentence. The central difficulty is the identification of a level of local evidence that already supports a candidate with foundational content and with a genuinely new organizing mechanism. Existing human workflows depend strongly on expert judgment, which makes large scale, parallel, and tool integrated discovery difficult.

Recent progress in language models for mathematical reasoning and formal proof has shown that models can participate in mathematically demanding tasks. Representative milestones include the quantitative reasoning system of \citet{lewkowycz2022solving}, the Lean 4 specialization route of \citet{wang2024theoremllama}, the natural language to formal proving system of \citet{yu2025mathesis}, and the olympiad level formal reasoning system of \citet{hubert2026olympiad}. Closely related formal systems from the last two years include the theorem proving systems of \citet{xin2024deepseekprover} and \citet{xin2024deepseekproverv15}, the interleaved thinking and proving framework of \citet{lin2024leanstar}, and the lifelong learning system of \citet{kumarappan2024leanagent}. Direct prompting of the form generate a new conjecture still does not reliably reach major conjecture level candidates. The common failure modes are rephrasing of famous templates, natural extensions of existing theory, and grand sounding narratives without a new proxy object or a structurally checkable claim. Such outputs usually lack the long term value required by high problem taste. The appropriate target is therefore not a single prompt, but a closed pipeline that covers candidate region search, high level semantic assessment, and formal inspection.

Major conjecture candidates must pass through validation layers with different roles. Search should start from regions that already exhibit a pre major conjecture state rather than from unconstrained enumeration over all mathematical space. Candidate statements must be evaluated for foundationality, novelty, and potential significance, otherwise formal novelty can replace mechanistic novelty. Natural language candidates must also be compressed into theorem statements that can enter a formal system. The mismatch between natural language and formal statements has been highlighted by \citet{ospanov2025minif2f} and \citet{yu2025mathesis}, and a larger Lean 4 benchmark from \citet{yu2025formalmath} indicates that this mismatch is a systemic bottleneck rather than a marginal nuisance. Even after a candidate enters Lean, the signals produced by exact? and aesop remain only probes of formal absorption and triviality, rather than terminal mathematical judgments. In the present framework, non-closure by exact? is a positive signal that the candidate is not immediately absorbed by the current library, and non-closure by aesop is a positive signal that the current formulation retains nontrivial structure. Major conjecture discovery is therefore closer to a heterogeneous validation chain than to a single model generation task.

Existing work is distributed across three adjacent areas, namely formal theorem proving, autoformalization, and AI assisted mathematical exploration. The environment of \citet{yang2023leandojo} focuses on proof search and premise selection after a formal statement is already available. High quality autoformalization work such as that of \citet{chan2025leaning} and \citet{yu2025mathesis} focuses on the quality of the mapping from natural language to formal language. Open ended mathematical exploration is represented by \citet{georgiev2025exploration}. Even so, the full problem of proposing, screening, and formalizing conjectures with major potential is still not addressed as a single workflow. The missing component is an explicit mechanism that biases search towards high problem taste. More precisely, most automated research systems optimize benchmark improvement, review scores, publication oriented clarity, or hypothesis generation under externally provided objectives. Those targets are not equivalent to the discovery of problems that would offer long term help to human mathematics after proof. This point is visible in recent research automation systems of \citet{baek2024researchagent}, \citet{lu2024aiscientist}, \citet{yamada2025aiscientistv2}, \citet{gottweis2025towards}, and Analemma \citeyearpar{analemma2025fars}. The same pattern appears in evaluation frameworks from \citet{chen2024scienceagentbench}, \citet{chen2025autobench}, \citet{wijk2024rebench}, and \citet{nathani2025mlgym}, which emphasize execution quality on bounded research tasks rather than problem taste in the sense of long term knowledge organization.

We address this gap with a three stage framework. The search stage starts from mathematical regions that already exhibit explicit local evidence modules. The reflective validation stage evaluates candidates in terms of foundationality, novelty, and potential significance. The formal validation stage compresses each candidate into a Lean statement and subjects it to checks of syntactic validity, library level absorption, and automatic triviality. The central ambition is not only the identification of candidates that are formally checkable, but also the more stable approximation of mathematical problems with high problem taste. The value of the framework lies in the transformation of major conjecture discovery from a single natural language act into a candidate management process that can be contracted, compared, audited, and inserted into a formal system.

The contributions of the paper are as follows.
\begin{itemize}
\item We formulate major conjecture discovery as a structured problem with explicit regional input, candidate level semantic representation, and formal validation records.
\item We introduce a three stage pipeline that combines local evidence driven search, reflective semantic validation, and formal validation in Lean 4 and Mathlib.
\item We define a validation protocol that separates syntactic validity, direct library absorption, and automatic triviality, so that formal inspection becomes a source of comparable structural signals rather than a binary proof outcome.
\item We provide experimental evidence on twenty candidates showing that the pipeline preferentially retains structurally dense candidates with higher semantic value and with stronger potential to provide durable help to human mathematical research.
\end{itemize}

\section{Related Work}
\label{sec_related}

\subsection{Formalization and theorem proving}
\label{sec_related_formal}

During the last two years, research on language models in Lean and other formal proof environments has moved rapidly from small scale zero shot trials to specialized data, specialized models, and end to end evaluation. \citet{wang2024theoremllama} illustrate a route from general purpose language models to Lean 4 experts. \citet{xin2024deepseekprover} and \citet{xin2024deepseekproverv15} show the importance of large synthetic corpora, proof assistant feedback, reinforcement learning, and search. \citet{lin2024leanstar} and \citet{kumarappan2024leanagent} further show that theorem proving can benefit from explicit intermediate reasoning traces and from cross repository continual learning. Work on autoformalization quality from \citet{chan2025leaning}, \citet{yu2025mathesis}, \citet{jana2025proofbridge}, and \citet{huang2025formarl} demonstrates that the quality of formalization is often the dominant factor behind overall theorem proving performance. The environment of \citet{yang2023leandojo} provides an important reproducible infrastructure for premise retrieval and theorem proving. The focus of this line of work is the question of how to formalize or prove a given problem. The present paper instead asks which candidates deserve formalization and why.

\subsection{Conjecture generation and automated research}
\label{sec_related_discovery}

The line of work closest to the present objective is automatic conjecture generation and AI assisted mathematical discovery. \citet{onda2025leanconjecturer} show that large scale generation of Lean 4 conjectures with syntax and nontriviality screening is feasible. Mathematical exploration at scale from \citet{georgiev2025exploration} demonstrates the power of open search over mathematical constructions. Related math AI work from \citet{wagner2021constructions}, \citet{charton2024patternboost}, and \citet{swirszcz2025geometry} shows that machine learning can also produce counterexamples, combinatorial constructions, and extremal polytopes. These contributions are directly relevant because they demonstrate that automated systems can discover mathematically valuable objects, while the optimization target remains explicit object construction rather than high taste problem selection. The olympiad level formal reasoning system of \citet{hubert2026olympiad} shows that formal mathematical reasoning can now enter substantially more difficult tasks. In the broader context of automated research, \citet{baek2024researchagent}, \citet{lu2024aiscientist}, \citet{yamada2025aiscientistv2}, \citet{gottweis2025towards}, and Analemma \citeyearpar{analemma2025fars} have all demonstrated partial automation of literature review, hypothesis generation, experiment execution, or manuscript preparation. The principal evaluation targets in these systems remain novelty of ideas, execution quality, paper quality, review score, bounded acceptance criteria, or task performance under a specified scientific goal. Those targets do not directly optimize the discovery of mathematically high taste problems. The same limitation appears in recent benchmarks for scientific agents from \citet{chen2024scienceagentbench}, \citet{chen2025autobench}, \citet{wijk2024rebench}, and \citet{nathani2025mlgym}. Related systems for scientific model discovery such as \citet{li2024automated} and \citet{shojaee2024llmsr} also confirm the effectiveness of proposal and critique loops over open model spaces, but the evaluation target remains model discovery quality rather than the taste of research problems themselves. The present work differs from these lines by treating high problem taste as an explicit methodological target and by integrating search, semantic selection, and formal inspection into a single chain.

\section{Method and Problem Formulation}
\label{sec_method}

\subsection{Structured input and structured output}
\label{sec_problem}

The discovery of major mathematical conjectures is treated as a structured computational problem rather than as a mapping from one natural language paragraph to another. The input is a mathematical region description \(R\), where the region is not simply the name of a field, but a composite object that contains local evidence and background information. We write
\begin{equation}
R = (\mathcal{D}, \mathcal{E}, \mathcal{P}, \mathcal{S}),
\end{equation}
where \(\mathcal{D}\) denotes a natural language description of the target mathematical region, \(\mathcal{E}\) denotes local evidence modules already present in that region, \(\mathcal{P}\) denotes known weak results, partial results, or background theoretical resources, and \(\mathcal{S}\) denotes an optional pool of candidate statements or candidate documents. The set \(\mathcal{S}\) can be empty, in which case search is launched directly from the regional description. When external candidate documents are available, \(\mathcal{S}\) serves as an input source for downstream validation.

The output is not a single best conjecture. The output is a candidate set
\begin{equation}
C = \{c_1, c_2, \dots, c_n\}.
\end{equation}
Each candidate \(c_i\) is represented as a joint record
\begin{equation}
c_i = (c_i^{\mathrm{nl}}, c_i^{\mathrm{sem}}, c_i^{\mathrm{formal}}, c_i^{\mathrm{meta}}),
\end{equation}
where \(c_i^{\mathrm{nl}}\) is the natural language statement, \(c_i^{\mathrm{sem}}\) is the structural semantic object that records object class, proxy object, invariants, relation type, and template, \(c_i^{\mathrm{formal}}\) is the Lean theorem statement together with machine validation results, and \(c_i^{\mathrm{meta}}\) records scores, risk flags, annotations, and provenance. This output design serves two direct purposes. First, the system does not need to enforce a single conjecture after one pass, and can preserve multiple candidates that survive into different validation layers. Second, candidates can be revised, compared, or discarded at different levels without being frozen at the natural language stage.

Under this formulation, the target is not conventional accuracy or proof success rate. The more appropriate objective is to construct, for a given region \(R\), a candidate set \(C\) whose elements satisfy three criteria as strongly as possible. The first criterion is structural expressibility, namely whether a candidate can be compressed into a clear object, invariant, relation structure and then written as a Lean statement. The second criterion is semantic value, namely the joint performance of the candidate on foundationality, novelty, and potential significance. The third criterion is formal checkability, namely whether the candidate produces meaningful signals in Lean and Mathlib rather than collapsing immediately into a known theorem, a trivial proposition, or a semantically unclear pseudo target.

\subsection{Pipeline overview}
\label{sec_pipeline}

\begin{figure*}[t]
\centering
\includegraphics[width=0.95\textwidth]{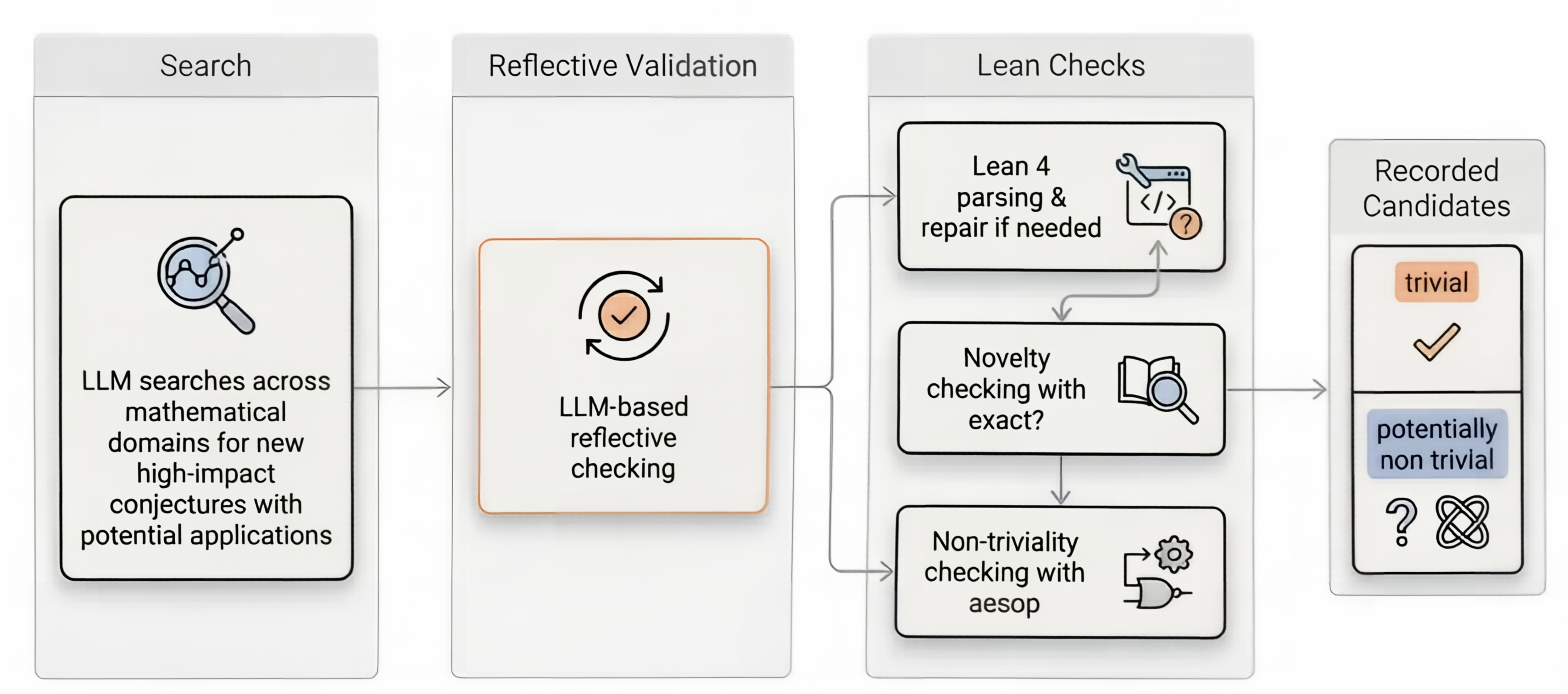}
\caption{Overview of the three stage pipeline. The search stage proposes high potential candidates from mathematical regions with explicit local evidence modules. The reflective validation stage performs high level screening in terms of foundationality, novelty, and potential significance. The Lean stage performs parsing, novelty checks with exact?, and nontriviality checks with aesop. The output is not a binary proof result. The output is a collection of formally recorded candidates with comparable structural signals.}
\label{fig_pipeline}
\end{figure*}

The overall pipeline is shown in \Cref{fig_pipeline}. Rather than using a single generator, we formulate the method as a multi stage contraction process. Given a region description \(R\), the candidate search module produces an initial set \(C^{(0)}\), the reflective validation module contracts that set to \(C^{(1)}\), and the Lean validation module produces a final collection of formal validation records \(C^{(2)}\). The pipeline can therefore be written as
{\setlength{\abovedisplayskip}{4pt}
\setlength{\abovedisplayshortskip}{4pt}
\setlength{\belowdisplayskip}{4pt}
\setlength{\belowdisplayshortskip}{4pt}
\begin{equation}
R \;\longmapsto\; C^{(0)} \;\longmapsto\; C^{(1)} \;\longmapsto\; C^{(2)}.
\end{equation}
}
As the pipeline advances, the size of the candidate set decreases, while the structural rigidity and validation depth of the candidates increase. The central difference across stages is not the identity of the model, but the knowledge function assigned to each stage. The first stage is responsible for candidate space coverage. The second stage is responsible for semantic density. The third stage is responsible for machine checkability.

This formulation differs sharply from standard theorem proving workflows. The systems of \citet{yang2023leandojo} and \citet{wang2024theoremllama} typically take an external theorem statement as the starting point, and then optimize proof success or premise retrieval. The present problem begins before a theorem statement exists. The central questions are how a candidate is proposed, how it is compared to alternatives, and how it is compressed into a statement that can enter a formal system. As a result, final proof success is not treated as the only objective. The ability of a candidate to survive and remain meaningful across stages is treated as equally important. Failure to move from search to reflective validation indicates a lack of structural support. Failure to move from reflective validation to Lean indicates that the statement remains too vague. Immediate closure by library results or automatic tactics indicates that the formal expression lacks genuine research tension.

\subsection{Candidate search from local evidence modules}
\label{sec_search}

The objective of the first stage is not to output a terminal major conjecture immediately. The objective is to identify mathematical regions with genuine potential for major conjecture formation, and to formulate structured candidate drafts inside those regions. We use a region driven search strategy rather than a proposition driven strategy because major conjectures do not appear uniformly in mathematical space. They tend to emerge in regions that already exhibit recognizable evidence patterns. Such regions often share several features. Stable empirical regularities appear before complete structural explanations. Main terms or boundary cases already exist. Weak versions form a chain of approximation. Local obstacles can be stated explicitly. A high fidelity proxy object begins to emerge. These signals are abstracted into local evidence modules and used as structural priors for search.

The search function does not determine whether a proposition is correct. The function determines whether a region has reached a pre major conjecture state. The score depends on the joint configuration of evidence modules rather than on a single statistic. Relevant signals include the stability of empirical regularities, the continuity of the weak result chain, the explicitness of local obstructions, the fidelity of available proxy objects, and the presence of tensions of the form main term known while limiting error remains unresolved. Region level filtering avoids blind enumeration over the full mathematical space and reduces the probability that a language model falls back to familiar training corpus templates.

The output of the search stage is a candidate draft represented by a structured record
\begin{equation}
v_i^{(0)} = \big(\mathrm{domain}(c_i), \mathrm{proxy}(c_i), \mathrm{relation}(c_i), \mathrm{evidence}(c_i)\big).
\end{equation}
The domain field records the mathematical area, the proxy field records the proposed high fidelity object, the relation field records the tentative relation type, and the evidence field preserves the local evidence chain. Compared with the direct large scale conjecture generation strategy of \citet{onda2025leanconjecturer}, the present stage emphasizes interpretability of provenance rather than raw conjecture count.

\subsection{Reflective semantic validation}
\label{sec_reflective}

The candidates produced by the first stage are only candidate drafts. The second stage therefore performs high level semantic validation before proof or full formalization is attempted. We use three evaluation dimensions, namely foundationality, novelty, and potential significance. The point of making these dimensions explicit is that the value of a major conjecture is not guaranteed by its surface form. Value depends on the place of the candidate within the mathematical knowledge network, on whether the proposed mechanism is genuinely new, and on whether a proof could reorganize a research region or spill over into logic, computer science, cryptography, physics, or other mathematical areas.

Foundationality asks whether the candidate addresses root level structure rather than a local estimate or surface variation. Novelty asks whether the candidate differs substantially from existing famous templates, natural extensions of existing theories, and common conjectural packaging. Potential significance does not require immediate engineering application. The criterion asks whether a proof could reorganize the language of a region, motivate new proxy objects, or generate methodological spillover. The reflective stage records the score vector
\begin{equation}
s(c_i) = \big(s_f(c_i), s_n(c_i), s_a(c_i)\big),
\end{equation}
where the three components denote foundationality, novelty, and potential significance respectively. In addition to scores, the stage records rationales and risk flags,
\begin{equation}
v_i^{(1)} = \big(s(c_i), \mathrm{rationale}(c_i), \mathrm{risk\_flags}(c_i)\big).
\end{equation}
These records make explicit why a candidate is retained and why another candidate is deferred. Many candidates fail not because they are obviously false, but because their object class remains unstable, their proxy object remains too vague, their application chain is too long, or their structure remains too close to established templates.

\subsection{Lean validation and formal signals}
\label{sec_formal}

After reflective validation, the third stage moves candidates into Lean 4 and Mathlib so that they can be subjected to machine checkable constraints. Lean is not used because it can already prove major conjectures directly. Lean and Mathlib are used because they provide a unified statement interface through which three questions can be treated programmatically, namely whether a candidate can be parsed, whether it is already absorbed by the current library, and whether the current expression is too weak. The formal system functions here as a structural probe rather than as a terminal judge.

The first signal is syntactic validity. For each candidate \(c_i\), the system constructs a Lean theorem statement \(\tau(c_i)\) and checks whether \(\tau(c_i)\) can be parsed and type checked in Lean 4. Syntactic validity does not imply mathematical truth, but it is the minimum requirement for entry into formal validation. If a candidate cannot be compressed into a clear theorem statement, then the candidate remains at the level of natural language intuition. The second signal is formal novelty. We do not ask whether the candidate has never appeared in mathematical history. Instead, we use exact? as a proxy for direct absorption by the current Mathlib environment. If exact? closes the goal directly, then the candidate is treated as directly solved in the current formal context. If exact? does not close the goal, then the candidate is treated as not directly solved, and this is interpreted as positive evidence that the candidate has not yet been absorbed by the current library layer. The third signal is nontriviality. We use aesop as a probe of automatic triviality at the current statement level. If aesop proves the theorem directly, the candidate is marked as trivial. If aesop does not prove the theorem, then the candidate is marked as potentially nontrivial, and this is interpreted as positive evidence that the current formulation still retains structural tension.

The final formal record is
\begin{equation}
v_i^{(2)} = \big(\mathrm{syntax}(c_i), \mathrm{novelty}_{\mathrm{Mathlib}}(c_i), \mathrm{nontriviality}(c_i)\big).
\end{equation}
This record distinguishes three classes of problems. Syntax failure usually indicates that the statement is unclear. Success of exact? usually indicates that the candidate remains too close to existing theory. Success of aesop usually indicates that the expression is too weak. Conversely, non-closure by exact? and non-closure by aesop are positive retention signals in the pipeline because they indicate that the candidate is neither immediately absorbed by the current library nor automatically trivial at the present statement level. The role of the third stage is therefore not to return a final verdict, but to provide uniform and comparable formal signals for candidate management.

\section{Experiments and Evaluation}
\label{sec_experiments}

\subsection{Experimental goals}
\label{sec_goals}

The experiments do not attempt to determine the mathematical truth of the candidates. The goal is to test whether the three stage pipeline can stably generate, review, and pass major conjecture candidates into a unified formal inspection environment. More specifically, the experiments test whether the search stage can produce candidates with explicit structural tension under strong prompt constraints, whether the reflective validation stage can provide discriminative signals for foundationality, novelty, and potential significance, and whether the formal stage can systematically map the candidates into Lean 4 and Mathlib and expose stable syntactic, novelty, and nontriviality signals. In short, the experiments do not ask only whether the system can generate new propositions. The experiments ask whether the system can generate candidates with high problem taste, namely candidates that could offer substantial assistance to human mathematical research if a proof became available.

\subsection{Candidate corpus and prompt materials}
\label{sec_corpus}

The experimental material has three parts, namely the candidate corpus, the prompt used for the first stage, and the prompt used for the second stage. The candidate corpus consists of the actual outputs of the first two stages and contains twenty candidates. Each candidate is paired with a natural language statement, a semantic scoring record from the reflective stage, and the structural information required by the formal stage. The experimental objects are therefore not arbitrary mathematical sentences collected post hoc. They form a unified corpus produced by GPT Pro under one constrained prompt policy and processed under one validation protocol. The candidate corpus is summarized in \Cref{tab_candidates}.

\begin{table*}[t]
\centering
\small
\caption{Candidate corpus and semantic scores. F denotes foundationality. N denotes novelty. P denotes potential significance.}
\label{tab_candidates}
\begin{tabular}{llrrr}
\toprule
ID & Title & F & N & P \\
\midrule
C001 & Missing Face Boundary Class Condensation Conjecture & 7.0 & 8.0 & 6.0 \\
C002 & Boundary Adelic Unification Conjecture for Random Complexes & 7.0 & 8.5 & 6.0 \\
C003 & Quantum Topological Holography Conjecture & 5.0 & 4.0 & 5.0 \\
C004 & Spectral Methods in Quantum Field Theory and Homology in Topological Data Analysis & 4.5 & 3.5 & 4.5 \\
C005 & Equivalence Between Geometric Features of Chaotic Systems and Algebraic Geometric Objects & 4.5 & 3.5 & 4.0 \\
C006 & Local Witness Transfer Resonance Conjecture & 8.0 & 8.0 & 8.0 \\
C007 & Adelic Smith Field Conjecture & 8.0 & 8.5 & 6.5 \\
C008 & Betti Foliation and Normal Function Rigidity Direction & 8.0 & 7.0 & 6.0 \\
C009 & Proxy Object and Error Control Framework & 3.5 & 3.0 & 4.0 \\
C010 & Spectral Discreteness Conjecture in Loop Space for Four Dimensional Pure Yang--Mills & 9.0 & 8.0 & 7.5 \\
C011 & Stationary Horizon and Busemann Web Rigidity in KPZ Geometry & 8.5 & 8.0 & 7.0 \\
C012 & Multi Direction Candidate Screening Draft & 2.0 & 2.0 & 2.0 \\
C013 & Boundary Extendability Spectral Unification Conjecture & 8.5 & 8.5 & 8.5 \\
C014 & Spectral Physical Phase Transition Conjecture in High Dimensional Random Constraint Systems & 4.5 & 3.5 & 5.5 \\
C015 & Correspondence Between Quantum Entanglement Distributions and Prime Distributions & 3.5 & 2.5 & 5.0 \\
C016 & Adaptive Fractal Geometry and Dynamical Systems Conjecture & 4.0 & 3.0 & 4.5 \\
C017 & Three Direction Conjecture List & 2.0 & 2.0 & 2.0 \\
C018 & Spectral Structure and Phase Transition Conjecture for High Dimensional Topological Data & 4.0 & 3.0 & 5.0 \\
C019 & Quantum Chaotic Geometry Conjecture & 4.0 & 3.0 & 4.5 \\
C020 & Conjecture on Boundary Layers and Global Solutions in High Dimensional Fluid Dynamics & 4.5 & 3.5 & 6.0 \\
\bottomrule
\end{tabular}
\end{table*}

\Cref{tab_candidates} shows that the twenty candidates are not distributed uniformly across one quality band. Candidates such as C006, C007, C010, C011, and C013 occupy a clearly higher region of the score space. Lower scoring candidates cluster around broader cross domain analogies that still lack sufficient support in foundationality or novelty. This pattern is important because it shows that the pipeline does not simply produce uniformly grand statements. Instead, the pipeline introduces separation inside a single batch of outputs and isolates a much smaller set of structurally dense candidates. In that sense, the system begins to approximate a search process for high taste problems rather than a generator of generic novelty.

The prompt itself is treated as part of the experimental material because the central methodological novelty lies not only in calling a language model, but also in encoding search constraints into the input. The first prompt constrains the search target and disallows migration of existing famous conjecture templates. The second prompt reorganizes the same model family into a semantic critic. This design is methodologically related to the self critique ideas of \citet{shinn2023reflexion} and \citet{madaan2023selfrefine}, but the purpose here is not generic output refinement. The purpose is to turn semantic retention and semantic rejection into explicit experimental records.

\subsection{Evaluation protocol}
\label{sec_protocol}

The evaluation protocol follows the three stage pipeline, but the final statistics are concentrated into three metrics, namely syntactic validity, formal novelty, and nontriviality. The first two stages produce candidate records and semantic scores. The third stage produces unified machine validation signals. This design allows the experiments to cover the full path from natural language to formal language without collapsing all validation layers into one opaque scalar score.

The first stage is evaluated in terms of whether candidate generation produces objects that can proceed to the next stage. The primary concern is not grammatical pass or fail in isolation. The concern is whether the candidate already contains a usable object class, proxy object, invariant family, relation type, and statement template. The second stage is evaluated using the three score dimensions. For each candidate, the system records three scores on a ten point scale together with textual rationales. The goal is to ensure that candidates entering Lean already possess explicit theoretical positioning. The third stage is evaluated entirely in Lean 4 and Mathlib. Syntactic validity checks whether the theorem statement parses and type checks. Formal novelty checks whether exact? closes the theorem directly. Nontriviality checks whether aesop proves the theorem directly.

\subsection{Experimental environment}
\label{sec_environment}

The experiments are conducted in two execution environments with clearly separated roles. The first environment is used for candidate generation and reflective semantic review. In the present study, the search stage and the reflective validation stage are both executed with GPT Pro under the same constrained prompt materials. This stage does not depend on specialized proof infrastructure. The second environment is used for formal validation inside a clean Lean 4 and Mathlib project where all candidates are subjected to theorem statement parsing, exact? checks, and aesop checks. The purpose of this split is to distinguish high level semantic failure from formal failure. The Lean environment uses a stable Lean 4 and Mathlib combination inside one independent project, and all theorem statements are generated by a unified script. The resulting tactic signals therefore come from a real Mathlib environment rather than from a toy placeholder setting.

\subsection{Main results}
\label{sec_results}

At the current setting, the full three stage pipeline successfully passes all twenty candidates into a real Mathlib environment and produces unified formal signals for each one. The main quantitative summary appears in \Cref{tab_results}.

\begin{table}[t]
\centering
\caption{Main formal validation results}
\label{tab_results}
\begin{tabular}{ll}
\toprule
Metric & Result \\
\midrule
Syntactic Validity & 20/20 pass \\
Novelty with exact? & 20/20 not directly solved \\
Nontriviality with aesop & 20/20 not automatically trivial \\
Duplicate count & 0/20 \\
\bottomrule
\end{tabular}
\end{table}

\Cref{tab_results} establishes that the pipeline is executable end to end. Every candidate can be compressed into a Lean theorem statement and checked in the same environment. The result of twenty out of twenty syntax passes means more than a surface grammar success. It indicates that the transformation from natural language to formal theorem statements is stable at batch scale. The results for exact? and aesop must also be interpreted positively within the present framework. The important fact is not the presence of failure in an ordinary engineering sense. The important fact is that none of the candidates is immediately absorbed by the current library and none is discharged by a lightweight automatic tactic. At the current formal expression layer, these are positive indicators because they show that the candidate set preserves formal tension rather than collapsing into already available statements or automatic trivialities.

When the formal results are read together with the semantic score distribution in \Cref{tab_candidates}, a stronger conclusion becomes plausible. The pipeline does not simply generate novel looking sentences at random and then formalize them mechanically. The pipeline creates a layered filter through which only a smaller set of candidates retains both semantic density and formal tension. In other words, the pipeline begins to approximate a search process for high taste mathematical problems, where the expected benefit of a proof to human mathematics is substantially larger than the value of merely introducing a new statement.

\subsection{Representative case studies}
\label{sec_cases}

To understand the structural differences behind the aggregate statistics, we analyze three representative candidates, namely C013, C006, and C010. These three cases are not selected because they pass a single formal threshold that other candidates fail to pass. They are selected because they represent three different semantic types of major conjecture candidates that the present pipeline can retain, namely a highly combinatorial candidate, a local propagation candidate in proof complexity, and a spectral candidate at a continuous and discrete interface. Their comparison is summarized in \Cref{tab_cases}. The remaining seventeen cases are described in the appendix.

\begin{table*}[t]
\centering
\small
\caption{Comparison of the three representative cases}
\label{tab_cases}
\begin{tabularx}{\textwidth}{l>{\raggedright\arraybackslash}p{0.21\textwidth}>{\raggedright\arraybackslash}p{0.16\textwidth}>{\raggedright\arraybackslash}p{0.18\textwidth}c>{\raggedright\arraybackslash}p{0.22\textwidth}}
\toprule
ID & Title & Domain & Proxy object & Score & Formal signal \\
\midrule
C006 & Local Witness Transfer Resonance Conjecture & Proof complexity and SAT & Local witness transfer operator \(T_r(F)\) & 8.0 / 8.0 / 8.0 & syntax pass, exact? open, aesop open \\
C010 & Spectral Discreteness Conjecture in Loop Space for Four Dimensional Pure Yang--Mills & Yang--Mills theory & Renormalized loop space semigroup generator \(L_{\mathrm{loop}}\) & 9.0 / 8.0 / 7.5 & syntax pass, exact? open, aesop open \\
C013 & Boundary Extendability Spectral Unification Conjecture & Random CSP and average case proof complexity & Boundary extendability relation and renormalization operator & 8.5 / 8.5 / 8.5 & syntax pass, exact? open, aesop open \\
\bottomrule
\end{tabularx}
\end{table*}

\Cref{tab_cases} is useful not because it presents three isolated examples, but because it shows that one formal signal profile can still correspond to several different semantic types. C013 has the clearest object and threshold organization. C006 isolates a local propagation mechanism inside proof complexity. C010 illustrates a high level spectral interface that remains expressible through a finite proxy. The case studies therefore complement the aggregate statistics by showing how the same validation chain applies to different kinds of mathematically meaningful candidates.

\subsubsection{Case C013}
\label{sec_case_c013}

C013 concerns a stronger unification claim rather than a single threshold statement. For a finite domain random CSP template, the extendability of boundary assignments into a local neighborhood is organized into a boundary extendability relation distribution. When this distribution is written as a renormalization operator, the relative positions of the algorithmic threshold, the proof complexity threshold, and the satisfiability threshold become projections of one spectral bifurcation mechanism. The value of the case lies in the compression of threshold phenomena from algorithmics, probability, and proof complexity into one proxy object. The novelty does not arise merely from another threshold claim. The novelty arises from the introduction of the boundary extendability relation and its renormalization operator as the central organizing language.

The potential significance of C013 is substantial. If the conjecture is eventually validated, the organization of random CSP, SAT threshold theory, and average case proof complexity would likely change because algorithmic failure, proof difficulty, and phase diagram bifurcation would be read through the same local object. For example, the construction of hard SAT benchmarks would no longer need to depend mainly on heuristic random sampling. A researcher could instead search for instance families in which early bifurcation appears in the boundary extendability table while global satisfiability has not yet collapsed. Such a development would create a principled route from local boundary data to global complexity signals and would therefore provide direct assistance to human researchers who currently have to compare several disconnected threshold indicators.

\paragraph{Natural language statement.}
For every nondegenerate random CSP template \(X\) over a finite domain, if \(X\) is equipped with a boundary extendability relation distribution and if algorithmic thresholds, proof complexity thresholds, satisfiability thresholds, and a Lyapunov type invariant can all be observed on the same object, then the phase diagram of \(X\) is determined by spectral bifurcations of the corresponding renormalization operator.

\paragraph{Lean theorem statement.}
\begin{Verbatim}[fontsize=\small,breaklines=true]
def c013_prop : Prop :=
  forall X : Object,
    InClass "C013.object_class" X ->
    HasProxy "C013.proxy_object" X ->
    HasInvariant "C013.invariant.0" X ->
    HasInvariant "C013.invariant.1" X ->
    HasInvariant "C013.invariant.2" X ->
    HasInvariant "C013.invariant.3" X ->
    PhaseTransitionRel "C013.relation" X
\end{Verbatim}

\subsubsection{Case C006}
\label{sec_case_c006}

C006 is a local propagation candidate in proof complexity. The conjecture compresses several complexity quantities of explicit unsatisfiable 3-CNF families into one local operator \(T_r(F)\). The operator records how witnesses are transferred across the boundary of a local patch, and it induces a resonance radius \(R(F)\). The conjecture states that when \(R(F_n)\) has order \(\Theta(n)\), resolution width, PCR degree, resolution size, and CDCL time scales align. The value of the case lies in the attempt to rewrite a large family of proof complexity effects as consequences of one local propagation object. The novelty lies in the introduction of the witness transfer operator as a central proxy object rather than another reparameterization of width or degree.

The potential impact of C006 is unusually direct because the case touches SAT solving, proof logging, proof system comparison, and benchmark design. If the resonance radius does unify width, degree, size, and CDCL time, then one new variable would become available for identifying hard instances, diagnosing solver behaviour, and comparing different proof systems. A concrete example is current solver analysis, where learned clause statistics, restart behaviour, width proxies, and trace length are usually studied separately. Under C006, those observations could be compressed into one interpretable variable, which could then be used to determine whether an instance family is better matched to aggressive inprocessing, aggressive restart policies, or genuinely new clause learning designs. The main help to human researchers would be interpretive. Instead of reading many disconnected signals, researchers would gain one central quantity that explains why a system is difficult.

\paragraph{Natural language statement.}
For every explicit family \(X\) of unsatisfiable 3-CNF formulas, if \(X\) is equipped with a local witness transfer operator \(T_r(X)\) and if the resonance radius, resolution width, PCR degree, resolution size, and CDCL time can all be recorded on the same family, then the resonance radius \(R(X_n)\) has order \(\Theta(n)\) and aligns with the associated proof complexity measures and CDCL time scales.

\paragraph{Lean theorem statement.}
\begin{Verbatim}[fontsize=\small,breaklines=true]
def c006_prop : Prop :=
  forall X : Object,
    InClass "C006.object_class" X ->
    HasProxy "C006.proxy_object" X ->
    HasInvariant "C006.invariant.0" X ->
    HasInvariant "C006.invariant.1" X ->
    HasInvariant "C006.invariant.2" X ->
    HasInvariant "C006.invariant.3" X ->
    HasInvariant "C006.invariant.4" X ->
    AsymptoticRateRel "C006.relation" X
\end{Verbatim}

\subsubsection{Case C010}
\label{sec_case_c010}

C010 is a spectral candidate that lives at the interface between continuous Yang--Mills objects and finite proxies. The conjecture does not simply restate the mass gap problem. The conjecture introduces a new loop space proxy, namely a renormalized loop space semigroup generator \(L_{\mathrm{loop}}\) constructed from Wilson loop related quantities. The claim is that the area law, the mass gap, and long range loop factorization are different projections of one discrete pure point spectral structure. The value of the case lies in replacing three largely parallel phenomena by one spectral object. The novelty lies in the choice of \(L_{\mathrm{loop}}\) as the organizing proxy rather than local field variables or individual Wilson loop expectations.

The potential significance of C010 begins with the place of the underlying problem. Four dimensional pure Yang--Mills theory and the mass gap problem remain central open structures in mathematical physics. The proposed loop space spectral object could provide a common language across lattice gauge theory, loop equations, positivity bootstrap constraints, and finite truncation models. A concrete use case appears when two different loop basis truncations produce different low spectrum behaviour. At present it is difficult to tell whether an observed spectral gap is a numerical artefact or a genuine signal of the continuum limit. Under the proposed organization, stable appearance of isolated low spectrum together with matching Wilson area law decay would become a more coherent diagnostic criterion. The help to human researchers would therefore be methodological. The conjecture offers a possible path for aligning evidence that is currently distributed across several partially incompatible approximation schemes.

\paragraph{Natural language statement.}
For every continuum limit object \(X\) of four dimensional pure Yang--Mills theory over a compact simple Lie group, if \(X\) is equipped with a renormalized loop space semigroup generator \(L_{\mathrm{loop}}\) and if pure point spectrum, mass gap, Wilson loop area law decay, and multi loop factorization can all be recorded on the same object, then the area law, the mass gap, and long range loop factorization are jointly determined by one discrete loop space spectral structure.

\paragraph{Lean theorem statement.}
\begin{Verbatim}[fontsize=\small,breaklines=true]
def c010_prop : Prop :=
  forall X : Object,
    InClass "C010.object_class" X ->
    HasProxy "C010.proxy_object" X ->
    HasInvariant "C010.invariant.0" X ->
    HasInvariant "C010.invariant.1" X ->
    HasInvariant "C010.invariant.2" X ->
    HasInvariant "C010.invariant.3" X ->
    SpectralDiscretenessRel "C010.relation" X
\end{Verbatim}

The three cases show that the present method does not collapse all candidates into one stylistic family. The pipeline can retain combinatorial candidates, local propagation candidates, and spectral interface candidates while assigning the same formal signal profile to all three. More importantly, the difference among these cases is not cosmetic. The difference determines the kind of help a proof could provide to human mathematics, because different proxy objects reorganize different research workflows, evidence structures, and future proof strategies.

\section{Conclusion}
\label{sec_conclusion}

We have presented a three stage framework for the discovery of major mathematical conjectures through candidate region search, reflective semantic validation, and Lean based formal checks. Rather than treating major conjecture discovery as a one step prompting task or as a standard theorem proving task, the paper formulates the problem as a heterogeneous validation chain in which candidates must survive structural, semantic, and formal constraints.

Experiments on twenty candidates show that the full pipeline is executable end to end. Every candidate can be compressed into a Lean theorem statement in a real Lean 4 and Mathlib environment, and none of the twenty candidates is closed directly by exact? or proved directly by aesop. At the same time, the semantic score distribution and the case analysis preserve substantial distinctions among candidates. This indicates that the pipeline does not merely generate novelty. Instead, the pipeline provides a uniform comparison framework through which structurally denser and mathematically more consequential candidates can be separated from weaker outputs. The experimental evidence therefore supports a stronger claim. The pipeline is beginning to approximate a systematic search process for high taste mathematical problems, namely problems that could provide durable help to human mathematical research after proof.

The representative case analysis shows that one formal signal profile can still correspond to several distinct semantic types. Combinatorial threshold problems, local propagation problems in proof complexity, and spectral interface problems in mathematical physics can all survive the same validation chain while retaining their own object structures and impact patterns. The most important conclusion of the paper is therefore not that a major conjecture has already been proved automatically. The more important conclusion is that the discovery of major conjectures can now be reorganized as a process that is comparable, auditable, and compatible with formal mathematical systems. Under this perspective, the present work offers a methodological framework for the participation of large language models in open ended mathematical exploration, with explicit emphasis on the discovery of problems whose eventual proofs would most likely provide large and lasting benefits to human mathematics.

\section*{Impact Statement}
The present work studies a pipeline for the discovery and validation of mathematical conjectures. The intended contribution is methodological and is aimed at improving the quality, traceability, and formal checkability of mathematical idea generation. The system is not designed to replace mathematicians, nor does it claim to establish the truth of the proposed conjectures. The main broader impact lies in the possibility of assisting researchers in the identification of mathematically meaningful high quality problems and in the organization of supporting evidence. Potential risks arise from overstating the significance of automatically generated conjectures or from confusing formal checkability with mathematical truth. For that reason, the paper consistently treats the proposed outputs as candidates that require further mathematical scrutiny.

\bibliography{references}
\bibliographystyle{icml2025}

\newpage
\appendix
\onecolumn
\section{Additional Case Descriptions}
\label{app_cases}

This appendix provides English translations of the seventeen non-primary cases that do not appear in the main body. The purpose of the appendix is to record the concrete content, the significance and novelty, the potential impact and application significance, and the corresponding natural language and Lean level statements for each candidate. The formal result of twenty out of twenty syntax passes reported in the main body still refers to the unified formalization layer generated in the same Lean project.

For reference, \Cref{tab_appendix_all} lists all twenty candidates together with their domains and proxy objects, and \Cref{tab_appendix_other} lists the seventeen candidates that are not treated as primary cases in the main body.

\begin{table}[h]
\centering
\small
\caption{All twenty candidates}
\label{tab_appendix_all}
\begin{tabularx}{\textwidth}{l>{\raggedright\arraybackslash}X>{\raggedright\arraybackslash}p{0.22\textwidth}>{\raggedright\arraybackslash}p{0.23\textwidth}}
\toprule
ID & Title & Domain & Proxy object \\
\midrule
C001 & Missing Face Boundary Class Condensation & Random high dimensional topology & Missing top face boundary class cloud \\
C002 & Boundary Adelic Unification for Random Complexes & Random simplicial complexes and integral homology & Boundary adelic profile \\
C003 & Quantum Topological Holography & Quantum computing and topological data analysis & Quantum topological holographic map \\
C004 & Spectral Methods in Quantum Field Theory and Homology in TDA & Quantum field theory and topological data analysis & QFT spectral proxy \\
C005 & Chaotic Geometry and Algebraic Geometric Equivalence & Dynamical systems and algebraic geometry & Algebraic geometric proxy \\
C006 & Local Witness Transfer Resonance & Proof complexity and SAT & Local witness transfer operator \\
C007 & Adelic Smith Field & Random complexes and integral torsion & Prime indexed local Smith field \\
C008 & Betti Foliation and Normal Function Rigidity & Hodge theory and arithmetic geometry & Betti foliation and Betti strata \\
C009 & Proxy Object and Error Control Framework & High dimensional systems & Stability or error control proxy \\
C010 & Loop Space Spectral Discreteness in Four Dimensional Pure Yang--Mills & Yang--Mills theory & Renormalized loop space generator \\
C011 & Stationary Horizon and Busemann Web Rigidity in KPZ Geometry & KPZ directed random geometry & Stationary horizon and Busemann web \\
C012 & Multi Direction Screening Draft & Mixed geometry, QFT, and topology & Not selected \\
C013 & Boundary Extendability Spectral Unification & Random CSP and average case proof complexity & Boundary extendability relation \\
C014 & Spectral Physical Phase Transition in High Dimensional Random Constraint Systems & High dimensional random constraint systems & Spectral physical proxy \\
C015 & Quantum Entanglement and Prime Distribution Correspondence & Quantum computing and number theory & Entanglement distribution proxy \\
C016 & Adaptive Fractal Geometry and Dynamical Systems & Fractal geometry and dynamical systems & Time varying fractal dimension \\
C017 & Three Direction Conjecture List & Arithmetic geometry, random matrices, and dynamical number theory & Not selected \\
C018 & Spectral Structure and Phase Transition for High Dimensional Topological Data & Topological data analysis & Spectral structure proxy \\
C019 & Quantum Chaotic Geometry & Quantum geometry and chaotic dynamics & Quantum chaotic geometry \\
C020 & Boundary Layers and Global Solutions in High Dimensional Fluid Dynamics & Fluid dynamics & High dimensional vortex and boundary proxy \\
\bottomrule
\end{tabularx}
\end{table}

\begin{table}[h]
\centering
\small
\caption{Appendix cases with semantic scores}
\label{tab_appendix_other}
\begin{tabularx}{\textwidth}{l>{\raggedright\arraybackslash}Xrrrl}
\toprule
ID & Title & F & N & A & In main body \\
\midrule
C001 & Missing Face Boundary Class Condensation & 7.0 & 8.0 & 6.0 & No \\
C002 & Boundary Adelic Unification for Random Complexes & 7.0 & 8.5 & 6.0 & No \\
C003 & Quantum Topological Holography & 5.0 & 4.0 & 5.0 & No \\
C004 & Spectral Methods in Quantum Field Theory and Homology in TDA & 4.5 & 3.5 & 4.5 & No \\
C005 & Chaotic Geometry and Algebraic Geometric Equivalence & 4.5 & 3.5 & 4.0 & No \\
C007 & Adelic Smith Field & 8.0 & 8.5 & 6.5 & No \\
C008 & Betti Foliation and Normal Function Rigidity & 8.0 & 7.0 & 6.0 & No \\
C009 & Proxy Object and Error Control Framework & 3.5 & 3.0 & 4.0 & No \\
C011 & Stationary Horizon and Busemann Web Rigidity in KPZ Geometry & 8.5 & 8.0 & 7.0 & No \\
C012 & Multi Direction Candidate Screening Draft & 2.0 & 2.0 & 2.0 & No \\
C014 & Spectral Physical Phase Transition in High Dimensional Random Constraint Systems & 4.5 & 3.5 & 5.5 & No \\
C015 & Quantum Entanglement and Prime Distribution Correspondence & 3.5 & 2.5 & 5.0 & No \\
C016 & Adaptive Fractal Geometry and Dynamical Systems & 4.0 & 3.0 & 4.5 & No \\
C017 & Three Direction Conjecture List & 2.0 & 2.0 & 2.0 & No \\
C018 & Spectral Structure and Phase Transition for High Dimensional Topological Data & 4.0 & 3.0 & 5.0 & No \\
C019 & Quantum Chaotic Geometry & 4.0 & 3.0 & 4.5 & No \\
C020 & Boundary Layers and Global Solutions in High Dimensional Fluid Dynamics & 4.5 & 3.5 & 6.0 & No \\
\bottomrule
\end{tabularx}
\end{table}

\subsection{Case C001 Missing Face Boundary Class Condensation Conjecture}

\paragraph{Concrete content.}
The candidate is located in random high dimensional topology of Linial and Meshulam type. The central object is a cloud of boundary classes associated with missing top dimensional faces. For a random \(d\)-dimensional simplicial complex with a complete \((d-1)\)-skeleton, every missing \(d\)-face leaves a boundary class in the quotient defined by the boundary operator. The conjecture states that a macroscopic condensation of these classes occurs near the critical window, and that giant shadow growth, torsion burst, and an outlier in the Gram spectrum of the free part are different projections of that single condensation event.

\paragraph{Significance and novelty.}
The importance of this candidate lies in the replacement of three parallel signals by one boundary class object. The candidate is novel because the cloud of boundary classes is elevated into the organizing center of the problem, rather than being treated as a secondary descriptive layer around Betti counts, spectral statistics, or torsion measurements. The present limitation is that the route toward broader external impact remains indirect.

\paragraph{Potential impact and application significance.}
If this conjecture were established, the result would affect random topology, null models in topological data analysis, higher order network analysis, and the diagnosis of hidden modes in chain complex codes. The main practical benefit for researchers would be interpretive. Giant shadow growth, torsion burst, and spectral anomalies could be read on one explanatory chain instead of three disconnected tracks. For example, if a simplicial complex generated from high dimensional data exhibits rapid shadow growth together with a spectral anomaly, the boundary class condensation language would offer a principled criterion for deciding whether the data have entered a genuinely new topological phase. The broader help to human mathematics would consist in a more unified language for organizing statistical phenomena that are currently studied in parallel.

\paragraph{Natural language statement.}
For every object \(X\) in the class of Linial and Meshulam type random \(d\)-complexes, if \(X\) carries a missing face boundary class cloud and if giant shadow density, torsion subgroup, and the Gram spectrum of the free part are all observable, then one condensation event jointly controls giant shadow growth, torsion burst, and the spectral outlier.

\paragraph{Lean simplified statement.}
\begin{Verbatim}[fontsize=\small,breaklines=true]
forall X : Object,
  InClass "C001.object_class" X ->
  HasProxy "C001.proxy_object" X ->
  HasInvariant "C001.invariant.0" X ->
  HasInvariant "C001.invariant.1" X ->
  HasInvariant "C001.invariant.2" X ->
  PhaseTransitionRel "C001.relation" X
\end{Verbatim}

Foundationality is 7.0. Novelty is 8.0. Potential significance is 6.0.

\subsection{Case C002 Boundary Adelic Unification Conjecture for Random Complexes}

\paragraph{Concrete content.}
This candidate concerns random simplicial complexes and integral homology. The proposed proxy object is a boundary adelic profile that combines the spectral edge at the Archimedean place with Smith normal form data across all primes \(p\). The conjecture states that torsion entropy, the edge of the smallest singular values, Sylow \(p\)-primary profiles, and deviations from Cohen and Lenstra type behaviour are all projections of one limiting adelic boundary profile in the critical window.

\paragraph{Significance and novelty.}
The significance lies in the explicit placement of real spectral information and \(p\)-adic torsion information inside one object. The novelty does not come from another statistic at one prime. The novelty comes from a framework that spans all primes and the real place simultaneously. The main unresolved issue is the lack of a fully explicit mathematical realization of the adelic profile.

\paragraph{Potential impact and application significance.}
Progress on this candidate would affect the theory of integral homology in random complexes, prime wise torsion statistics, interpretation of Cohen and Lenstra type phenomena, and the analysis of integral topological features in discrete data. The main benefit to researchers is comparative power. Two random complex families that look almost identical at the level of rational homology could still behave very differently at selected primes. The adelic profile would provide one object through which such differences can be analyzed systematically. A concrete example is a pair of models with similar Betti transitions but sharply different torsion profiles at \(p=2\) and \(p=3\). The adelic language could help determine whether the divergence comes from different prime projections of one boundary matrix or from a deeper difference in the models. The broader value for human mathematics lies in a more systematic explanation of why integral homology carries structural information beyond rational homology.

\paragraph{Natural language statement.}
For every object \(X\) in a natural sparse random \(d\)-complex family with a stable critical window, if \(X\) carries a boundary adelic profile and if torsion entropy, the small singular value edge, Sylow \(p\)-primary profiles, and Cohen and Lenstra deviations are all present, then all key torsion phenomena are projections of one deterministic limiting adelic boundary profile.

\paragraph{Lean simplified statement.}
\begin{Verbatim}[fontsize=\small,breaklines=true]
forall X : Object,
  InClass "C002.object_class" X ->
  HasProxy "C002.proxy_object" X ->
  HasInvariant "C002.invariant.0" X ->
  HasInvariant "C002.invariant.1" X ->
  HasInvariant "C002.invariant.2" X ->
  HasInvariant "C002.invariant.3" X ->
  AsymptoticRateRel "C002.relation" X
\end{Verbatim}

Foundationality is 7.0. Novelty is 8.5. Potential significance is 6.0.

\subsection{Case C003 Quantum Topological Holography Conjecture}

\paragraph{Concrete content.}
This candidate introduces a quantum topological holographic map between quantum systems and topological objects such as random complexes. The intended claim is that the topology of quantum state space and the homological invariants of the associated topological object are systematically related.

\paragraph{Significance and novelty.}
The candidate is significant because it proposes a bridge between quantum state geometry and topological data analysis. The novelty lies in placing quantum state topology and random complex homology on two sides of one proxy structure. The candidate remains immature because quantifiers, maps, and checkable relations are not yet mathematically explicit.

\paragraph{Potential impact and application significance.}
If a rigorous version became available, the candidate could affect the topological readout of quantum state spaces, structural compression in quantum information, and shape aware feature extraction in quantum machine learning. A concrete research use case would be the comparison of two variational quantum ansatz families with similar energy behaviour but very different optimization trajectories. A rigorous holographic map might reveal geometric differences through Betti numbers or homological summaries. The main help to human researchers would be the introduction of a new observational coordinate for quantum state space, where structure could be compared by shape rather than only by fidelity or entropy.

\paragraph{Natural language statement.}
For every quantum system object \(X\), if \(X\) carries a quantum topological holographic map and if homology groups, Betti numbers, and the topology of quantum state space are observable, then the topological invariants of a random complex correspond systematically to the topology of the quantum state space.

\paragraph{Lean simplified statement.}
\begin{Verbatim}[fontsize=\small,breaklines=true]
forall X : Object,
  InClass "C003.object_class" X ->
  HasProxy "C003.proxy_object" X ->
  HasInvariant "C003.invariant.0" X ->
  HasInvariant "C003.invariant.1" X ->
  HasInvariant "C003.invariant.2" X ->
  EquivalenceRel "C003.relation" X
\end{Verbatim}

Foundationality is 5.0. Novelty is 4.0. Potential significance is 5.0.

\subsection{Case C004 Spectral Methods in Quantum Field Theory and Homology in Topological Data Analysis}

\paragraph{Concrete content.}
This candidate proposes a direct relation between spectral objects in quantum field theory and homology groups in topological data analysis. The intended claim is not a loose association. The intended claim is that spectral edge data or related spectral structure should determine Betti numbers and homology groups.

\paragraph{Significance and novelty.}
The significance lies in the attempt to place spectral language from quantum field theory and topological language from topological data analysis inside one coordinate system. The novelty comes from a direct cross domain bridge. The main obstacle is that spectra and homology groups are not naturally objects of the same type without an explicit intermediate construction.

\paragraph{Potential impact and application significance.}
If a mathematically rigorous version were eventually developed, the candidate could affect high dimensional data analysis, physically informed spectral methods, and the representation of complex data geometry. A direct research use case appears in settings where topological computation is expensive and parameter sensitive. If a spectral proxy from quantum field theory could reliably predict homological structure, then spectral analysis could act as a prescreening tool before the full topological computation is launched. The benefit to human researchers would be the introduction of a front end diagnostic, through which hidden topological structure becomes partially visible before the expensive homological stage.

\paragraph{Natural language statement.}
For every object \(X\) formed by pairing a quantum field with a topological data object, if \(X\) carries a spectral proxy from quantum field theory and if field spectrum, homology groups, and Betti numbers are observable, then the field spectrum and the homology groups of the data object are structurally equivalent.

\paragraph{Lean simplified statement.}
\begin{Verbatim}[fontsize=\small,breaklines=true]
forall X : Object,
  InClass "C004.object_class" X ->
  HasProxy "C004.proxy_object" X ->
  HasInvariant "C004.invariant.0" X ->
  HasInvariant "C004.invariant.1" X ->
  HasInvariant "C004.invariant.2" X ->
  EquivalenceRel "C004.relation" X
\end{Verbatim}

Foundationality is 4.5. Novelty is 3.5. Potential significance is 4.5.

\subsection{Case C005 Equivalence Between Geometric Features of Chaotic Systems and Algebraic Geometric Objects}

\paragraph{Concrete content.}
The candidate proposes an algebraic geometric proxy for chaotic behaviour. Orbit topology, bifurcation structure, singularities, and algebraic geometric structure are to be placed in one equivalence relation. The conjecture is therefore stronger than the idea that dynamical systems can be described geometrically. The conjecture states that chaotic behaviour itself may be equivalent to a specific algebraic geometric structure.

\paragraph{Significance and novelty.}
The candidate is significant because long standing links between dynamical systems and algebraic geometry remain distributed across several technical literatures. The novelty lies in an attempt to move chaotic behaviour itself from an analytic and topological description into an algebraic geometric object. The main limitation is the absence of a fixed system class and the absence of an explicit functor or equivalence mechanism.

\paragraph{Potential impact and application significance.}
If such an encoding were possible, the result would affect the modelling of complex dynamical systems, the classification of bifurcation regimes, the analysis of singular structures, and perhaps the algebraic representation of selected physical systems. The practical importance lies in the search for a usable geometric or algebraic language for phenomena that are usually treated as numerically unpredictable. A concrete example would be two chaotic systems with similar Lyapunov exponents but clearly different orbit geometry. An algebraic geometric proxy could help distinguish same mechanism chaos from qualitatively different mechanism chaos more stably than direct numerical inspection. The main benefit to human researchers would be a move away from purely heuristic numerical comparison and toward a more transferable geometric language.

\paragraph{Natural language statement.}
For every chaotic dynamical system object \(X\), if \(X\) carries an algebraic geometric proxy object and if orbit topology, singularities, and algebraic variety structure are observable, then the chaotic behaviour of \(X\) is equivalent to an associated algebraic geometric structure.

\paragraph{Lean simplified statement.}
\begin{Verbatim}[fontsize=\small,breaklines=true]
forall X : Object,
  InClass "C005.object_class" X ->
  HasProxy "C005.proxy_object" X ->
  HasInvariant "C005.invariant.0" X ->
  HasInvariant "C005.invariant.1" X ->
  HasInvariant "C005.invariant.2" X ->
  EquivalenceRel "C005.relation" X
\end{Verbatim}

Foundationality is 4.5. Novelty is 3.5. Potential significance is 4.0.

\subsection{Case C007 Adelic Smith Field Conjecture}

\paragraph{Concrete content.}
The conjecture again concerns random complexes and integral torsion, but the emphasis is more local and more \(p\)-adic than in C002. The core object is a prime indexed Smith field built from local Smith elimination, local minors, and \(p\)-adic valuation distributions. The candidate states that \(p\)-primary empirical measures, torsion burst times, critical window point processes, and the failure of Cohen and Lenstra type factorization are determined by this local integer linear algebra field.

\paragraph{Significance and novelty.}
The candidate is significant because many integral topological phenomena are intrinsically prime wise. The novelty lies in the shift from a global adelic profile to a local \(p\)-adic elimination cascade. This move makes the mechanism more local and more concrete, even though the distinction from C002 still requires clarification.

\paragraph{Potential impact and application significance.}
The practical significance of this direction lies in the development of finer prime wise diagnostics for integral topological statistics. The conjecture could affect random model analysis based on integer boundary matrices, chain complex codes, and local integer elimination studies in discrete structures. A concrete scenario is a random complex model that exhibits frequent torsion spikes at \(p=2\) but not at other primes. The Smith field language could help determine whether the effect is a systematic local elimination cascade or merely a finite sample accident. The help to human researchers lies in the creation of a local explanatory framework for prime specific integer phenomena, where such phenomena can be compared and accumulated rather than observed in isolation.

\paragraph{Natural language statement.}
For every natural sparse random \(d\)-complex or filtration object \(X\), if \(X\) carries a prime indexed local Smith field and if \(p\)-primary empirical measures, torsion burst time, critical window point processes, and failure of Cohen and Lenstra factorization are observable, then torsion statistics and burst thresholds are determined by that adelic Smith field.

\paragraph{Lean simplified statement.}
\begin{Verbatim}[fontsize=\small,breaklines=true]
forall X : Object,
  InClass "C007.object_class" X ->
  HasProxy "C007.proxy_object" X ->
  HasInvariant "C007.invariant.0" X ->
  HasInvariant "C007.invariant.1" X ->
  HasInvariant "C007.invariant.2" X ->
  HasInvariant "C007.invariant.3" X ->
  PhaseTransitionRel "C007.relation" X
\end{Verbatim}

Foundationality is 8.0. Novelty is 8.5. Potential significance is 6.5.

\subsection{Case C008 Betti Foliation and Normal Function Rigidity}

\paragraph{Concrete content.}
This candidate lies at the interface of Hodge theory and arithmetic geometry. The central object is not the cycle itself, but admissible normal functions arising from homologically trivial algebraic cycles, together with the Betti foliation, Betti strata, and Betti rank geometry induced on the relative intermediate Jacobian. The intended claim is that algebraicity, torsion and non-torsion behaviour, height dominance, and exceptional loci are all organized by this Betti foliation geometry.

\paragraph{Significance and novelty.}
The significance is high because the candidate places Chow groups, normal functions, Mumford and Tate defects, and arithmetic heights inside one mother region. The novelty comes from shifting the center of gravity from classical Hodge loci or cycle classes to the geometric position of normal functions inside the Betti foliation. The main limitation is not weak content, but the lack of a single final theorem form.

\paragraph{Potential impact and application significance.}
If this direction were clarified further, the result would affect the organization of algebraic cycles, arithmetic heights, exceptional sets, and moduli geometry. The importance does not lie in short term engineering impact. The importance lies in the provision of one mother object for several central questions in algebraic and arithmetic geometry. A concrete change in research workflow is easy to imagine. Instead of asking separately whether a family exhibits height growth, exceptional points, or torsion behaviour, a researcher could first ask how the family sits inside the Betti foliation and then derive the other questions from that geometry. The potential benefit to human mathematical research is therefore a reordering of how evidence is collected and how hard problems are decomposed.

\paragraph{Natural language statement.}
For every object \(X\) formed by a family of homologically trivial algebraic cycles and admissible normal functions, if \(X\) carries Betti foliation and Betti strata on the relative intermediate Jacobian, and if Betti rank, mixed Mumford and Tate quotient defect, and height dominance are observable, then algebraicity, height dominance, and exceptional loci are jointly controlled by the Betti foliation geometry.

\paragraph{Lean simplified statement.}
\begin{Verbatim}[fontsize=\small,breaklines=true]
forall X : Object,
  InClass "C008.object_class" X ->
  HasProxy "C008.proxy_object" X ->
  HasInvariant "C008.invariant.0" X ->
  HasInvariant "C008.invariant.1" X ->
  HasInvariant "C008.invariant.2" X ->
  StructureDecompositionRel "C008.relation" X
\end{Verbatim}

Foundationality is 8.0. Novelty is 7.0. Potential significance is 6.0.

\subsection{Case C009 Proxy Object and Error Control Framework}

\paragraph{Concrete content.}
This candidate concerns a high level framework for high dimensional dynamical systems or models in mathematical physics. The core idea is that a new proxy object could translate local structure, boundary perturbation, discrete spectrum, or dissipative mechanism into global error bounds and long term stability estimates.

\paragraph{Significance and novelty.}
The significance is methodological because many complex systems are difficult precisely because local structure does not accumulate naturally into global error control. The novelty lies in the insistence that a new proxy object is required instead of a patchwork of local estimates. The main limitation is that the candidate remains a framework rather than a theorem.

\paragraph{Potential impact and application significance.}
If a concrete version were developed, the framework could affect simulation of high dimensional systems, numerical stability analysis for partial differential equations, error propagation in mathematical physics, and robustness diagnostics for complex systems. The practical importance is substantial because many application settings still lack one common error proxy. A realistic use case is a simulation of a high dimensional dissipative system in which local error amplification is visible, but the connection to global failure remains unclear. A suitable proxy object could function as a risk indicator that predicts whether the system is entering a dangerous regime before full instability becomes visible. The help to researchers would therefore be operational as well as conceptual.

\paragraph{Natural language statement.}
For every high dimensional system object \(X\), if \(X\) carries a stability or error control proxy and if an error term, a stability measure, and a fluctuation scale are observable, then local structure controls global error.

\paragraph{Lean simplified statement.}
\begin{Verbatim}[fontsize=\small,breaklines=true]
forall X : Object,
  InClass "C009.object_class" X ->
  HasProxy "C009.proxy_object" X ->
  HasInvariant "C009.invariant.0" X ->
  HasInvariant "C009.invariant.1" X ->
  HasInvariant "C009.invariant.2" X ->
  UpperBoundRel "C009.relation" X
\end{Verbatim}

Foundationality is 3.5. Novelty is 3.0. Potential significance is 4.0.

\subsection{Case C011 Stationary Horizon and Busemann Web Rigidity in KPZ Geometry}

\paragraph{Concrete content.}
This candidate lies in directed random geometry of KPZ type. The core object is the stationary horizon or Busemann web inside the directed landscape scaling limit. The intended claim is that semi-infinite geodesics, exceptional geodesic networks, and multi type invariant measures are not parallel phenomena, but are jointly organized by one stationary horizon that can even determine the reconstruction of the geodesic network.

\paragraph{Significance and novelty.}
The candidate is significant because it targets one of the most central continuous limit objects in KPZ geometry. The novelty does not lie in another universality claim. The novelty lies in placing the stationary horizon or Busemann web into the role of a unique organizing center. The limitation is that the final theorem form remains to be compressed.

\paragraph{Potential impact and application significance.}
The significance of this direction begins with the internal reorganization of KPZ theory, random growth, and geodesic network geometry. The spillover could reach interacting particle systems, random metric geometry, and the study of extreme path structures. A concrete example appears in current work on the directed landscape, where semi-infinite geodesics, exceptional networks, and invariant measures are often estimated separately. If the stationary horizon became the unique organizing core, then one could study the horizon first and derive the other structures from it. The potential help to human researchers would be a reduction of fragmentation in KPZ limit theory. Instead of many beautiful but parallel objects, the field would gain a smaller set of central objects that organize the rest.

\paragraph{Natural language statement.}
For every directed landscape or KPZ scaling limit object \(X\), if \(X\) carries a stationary horizon or Busemann web and if semi-infinite geodesics, exceptional geodesic networks, and multi type invariant measures are observable, then the stationary horizon uniquely controls the reconstruction of the geodesic network.

\paragraph{Lean simplified statement.}
\begin{Verbatim}[fontsize=\small,breaklines=true]
forall X : Object,
  InClass "C011.object_class" X ->
  HasProxy "C011.proxy_object" X ->
  HasInvariant "C011.invariant.0" X ->
  HasInvariant "C011.invariant.1" X ->
  HasInvariant "C011.invariant.2" X ->
  UniquenessRel "C011.relation" X
\end{Verbatim}

Foundationality is 8.5. Novelty is 8.0. Potential significance is 7.0.

\subsection{Case C012 Multi Direction Candidate Screening Draft}

\paragraph{Concrete content.}
This entry is not a single conjecture. The entry retains several candidate regions in high dimensional statistical physics, random geometry, nonperturbative quantum field theory, and high dimensional geometric limits. The object of record is not a theorem, but a structured set of possible directions that appear to satisfy at least some conditions for major conjecture formation.

\paragraph{Significance and novelty.}
The value lies in the explicit exposure of high value regions inside the search space. The novelty is at the level of region selection rather than theorem formulation. The main role of this entry is as a boundary case showing that the system can stop before premature overcommitment.

\paragraph{Potential impact and application significance.}
The importance of this case is methodological. It shows that a discovery system can preserve uncertainty without collapsing into arbitrary precision. A realistic use case is a research team or a downstream automated system that faces several directions that all look mathematically important. The output of a structured candidate cluster can serve as the next input for human triage instead of forcing an artificial best answer. The help to human researchers is therefore not the direct solution of a problem, but a better allocation of attention under open ended uncertainty.

\paragraph{Natural language statement.}
For every object \(X\) that belongs to one of several retained candidate regions, if \(X\) carries the retained proxy object and a small set of initial invariants, then \(X\) supports at most a generic structure decomposition relation and no unique final relation has been selected.

\paragraph{Lean simplified statement.}
\begin{Verbatim}[fontsize=\small,breaklines=true]
forall X : Object,
  InClass "C012.object_class" X ->
  HasProxy "C012.proxy_object" X ->
  HasInvariant "C012.invariant.0" X ->
  StructureDecompositionRel "C012.relation" X
\end{Verbatim}

Foundationality is 2.0. Novelty is 2.0. Potential significance is 2.0.

\subsection{Case C014 Spectral Physical Phase Transition in High Dimensional Random Constraint Systems}

\paragraph{Concrete content.}
This candidate proposes one spectral physical proxy object for high dimensional random constraint systems such as random tensors, random graphs, and random matrices. The intended claim is that the proxy determines phase points, critical temperature, entropy related quantities, and asymptotic behaviour.

\paragraph{Significance and novelty.}
The significance lies in the attempt to compress local constraints and global phase transitions into one object. The novelty lies in placing spectral data, thermodynamic quantities, and phase thresholds on the same structural carrier. The candidate remains too broad at present.

\paragraph{Potential impact and application significance.}
If a concrete version were obtained, the candidate could influence the analysis of high dimensional random systems, models of statistical physics, complex networks, and energy landscapes in high dimensional machine learning. The practical value lies in the prospect of a common phase indicator across models. For example, researchers currently design separate diagnostics for random graphs, tensors, and matrices. A proxy object that survives across all three would permit direct comparison of phase evidence that is currently incommensurable. The benefit to human research would be a stronger ability to transfer local experience from one model family into wider structural judgment.

\paragraph{Natural language statement.}
For every high dimensional random constraint system object \(X\), if \(X\) carries a spectral physical proxy and if spectral distribution, critical temperature, and entropy measure are observable, then the location of phase points and the asymptotic behaviour of \(X\) are determined by that proxy.

\paragraph{Lean simplified statement.}
\begin{Verbatim}[fontsize=\small,breaklines=true]
forall X : Object,
  InClass "C014.object_class" X ->
  HasProxy "C014.proxy_object" X ->
  HasInvariant "C014.invariant.0" X ->
  HasInvariant "C014.invariant.1" X ->
  HasInvariant "C014.invariant.2" X ->
  PhaseTransitionRel "C014.relation" X
\end{Verbatim}

Foundationality is 4.5. Novelty is 3.5. Potential significance is 5.5.

\subsection{Case C015 Correspondence Between Quantum Entanglement Distributions and Prime Distributions}

\paragraph{Concrete content.}
This candidate attempts to use the distribution of quantum entanglement states as a proxy for prime distribution. The ambition is stronger than the idea that quantum computation might help number theory. The candidate proposes a structural correspondence between entanglement distributions and prime statistics.

\paragraph{Significance and novelty.}
The significance lies in the search for a representation of number theoretic statistics that differs sharply from the standard language of analytic number theory. The novelty comes from the direct rewriting of prime structure into an entanglement distribution proxy. The main limitation is the absence of a constructive bridge between quantum states and arithmetic objects.

\paragraph{Potential impact and application significance.}
If a workable version existed, the conjecture could influence number theoretic readout in quantum algorithms, integer statistical models in quantum cryptography, and the more basic question of whether quantum states can encode arithmetic regularity in a mathematically informative way. A concrete example is the possibility that low dimensional summaries of entanglement distributions might be used to describe local irregularities in prime statistics, rather than only classical analytic quantities. The main value to researchers would be the opening of a representation format that does not currently exist, even if no immediate solution to prime problems follows.

\paragraph{Natural language statement.}
For every object \(X\) that jointly involves quantum entanglement states and prime structure, if \(X\) carries an entanglement distribution proxy for prime structure and if entanglement distributions and prime distributions are observable, then a structural correspondence exists between the two.

\paragraph{Lean simplified statement.}
\begin{Verbatim}[fontsize=\small,breaklines=true]
forall X : Object,
  InClass "C015.object_class" X ->
  HasProxy "C015.proxy_object" X ->
  HasInvariant "C015.invariant.0" X ->
  HasInvariant "C015.invariant.1" X ->
  EquivalenceRel "C015.relation" X
\end{Verbatim}

Foundationality is 3.5. Novelty is 2.5. Potential significance is 5.0.

\subsection{Case C016 Adaptive Fractal Geometry and Dynamical Systems}

\paragraph{Concrete content.}
This candidate concerns a coupling between fractal geometry and dynamical systems. The key idea is that the fractal dimension of a certain nonlinear system should evolve through time and should be governed by a dynamical law. Fractal dimension is therefore treated as a time dependent variable rather than as a static descriptor.

\paragraph{Significance and novelty.}
The significance lies in the fact that many difficult systems exhibit geometric complexity that changes over time, while classical fractal geometry often studies static objects. The novelty lies in the elevation of time varying dimension into a central invariant. The main limitation is that the object class, the evolution law, and the measurable notion of dimension remain open.

\paragraph{Potential impact and application significance.}
If made rigorous, the direction could influence the analysis of complex time series in climate science, finance, life science, and dissipative dynamics. The important point is the availability of one quantity through which changing complexity can be modeled coherently. A typical scenario is a system moving from a relatively stable regime into turbulence, collapse, or abnormal oscillation. Researchers often track Lyapunov exponents or spectra in such settings. A time varying fractal dimension could function as a portable complexity instrument across domains. The resulting help to human researchers would lie in the ability to compare nonlinear phenomena from different fields with one common geometric quantity.

\paragraph{Natural language statement.}
For every adaptive fractal object \(X\) inside a nonlinear dynamical system, if \(X\) carries a time varying fractal dimension and if dimension, temporal evolution, and a governing dynamical law are observable, then the variation of dimension is controlled by that law.

\paragraph{Lean simplified statement.}
\begin{Verbatim}[fontsize=\small,breaklines=true]
forall X : Object,
  InClass "C016.object_class" X ->
  HasProxy "C016.proxy_object" X ->
  HasInvariant "C016.invariant.0" X ->
  HasInvariant "C016.invariant.1" X ->
  HasInvariant "C016.invariant.2" X ->
  AsymptoticRateRel "C016.relation" X
\end{Verbatim}

Foundationality is 4.0. Novelty is 3.0. Potential significance is 4.5.

\subsection{Case C017 Three Direction Conjecture List}

\paragraph{Concrete content.}
This entry preserves three directions simultaneously, namely an interaction between high dimensional arithmetic geometry and homological topology, a link between random matrix spectra and arithmetic structure, and a high dimensional geometric treatment of integer sequence dynamics. The entry therefore records three independent regions that the system judged to have embryo level major conjecture potential.

\paragraph{Significance and novelty.}
The value is similar to that of C012 and lies mainly at the level of exposing the search frontier rather than formulating one theorem. The novelty lies in structured regional selection rather than in a final mathematical claim. The entry demonstrates that the pipeline can preserve several strong directions without forcing a false sense of precision.

\paragraph{Potential impact and application significance.}
The main significance again lies at the level of method. The entry shows that the system can preserve several high potential regions in parallel and thereby support the allocation of research attention. A research team that is genuinely uncertain whether arithmetic geometry, random matrices, or dynamical number theory offers the richest next direction can use such a structured shortlist as an input for human deliberation. The value for human researchers lies in a better organization of uncertainty in the early phase of open ended work.

\paragraph{Natural language statement.}
For every object \(X\) that belongs to one of the three retained directions, if \(X\) carries the associated proxy object and the initial invariants, then \(X\) supports at most a generic structure decomposition relation and no unique final relation has yet been selected.

\paragraph{Lean simplified statement.}
\begin{Verbatim}[fontsize=\small,breaklines=true]
forall X : Object,
  InClass "C017.object_class" X ->
  HasProxy "C017.proxy_object" X ->
  HasInvariant "C017.invariant.0" X ->
  HasInvariant "C017.invariant.1" X ->
  HasInvariant "C017.invariant.2" X ->
  StructureDecompositionRel "C017.relation" X
\end{Verbatim}

Foundationality is 2.0. Novelty is 2.0. Potential significance is 2.0.

\subsection{Case C018 Spectral Structure and Phase Transition for High Dimensional Topological Data}

\paragraph{Concrete content.}
This candidate introduces a spectral structure proxy inside high dimensional topological data analysis and claims that spectral distributions can reveal global phase transition behaviour in addition to local topological features. The intended target is a relation between spectral criticality and the phase transitions visible through persistent homology.

\paragraph{Significance and novelty.}
The significance lies in the attempt to move topological data analysis from a language of shape description to a language of structural phase transition. The novelty comes from treating phase transition as a central part of the proxy object rather than as a secondary interpretation. The current weakness is the lack of a fixed data model, spectral operator, and threshold definition.

\paragraph{Potential impact and application significance.}
If the direction were developed further, it could affect high dimensional data analysis, network phase transition detection, geometric representation in machine learning, and topological warning systems in complex monitoring tasks. The important shift is from the use of topological data analysis as a static descriptive tool to its use as a tool for identifying structural criticality. A concrete scenario arises in sensor networks, brain networks, or material experiments, where one may observe changes in persistent homology while remaining unsure whether those changes are merely descriptive or whether they signal an approaching transition. A spectral structure proxy that aligns topology with criticality would provide a more actionable warning signal. The help to human researchers would therefore lie in a reinterpretation of shape change as structural transition evidence.

\paragraph{Natural language statement.}
For every high dimensional topological data object \(X\), if \(X\) carries a spectral structure proxy and if spectral distribution, persistent topological features, and critical behaviour are observable, then the spectral structure of \(X\) reveals its global phase transition behaviour.

\paragraph{Lean simplified statement.}
\begin{Verbatim}[fontsize=\small,breaklines=true]
forall X : Object,
  InClass "C018.object_class" X ->
  HasProxy "C018.proxy_object" X ->
  HasInvariant "C018.invariant.0" X ->
  HasInvariant "C018.invariant.1" X ->
  HasInvariant "C018.invariant.2" X ->
  PhaseTransitionRel "C018.relation" X
\end{Verbatim}

Foundationality is 4.0. Novelty is 3.0. Potential significance is 5.0.

\subsection{Case C019 Quantum Chaotic Geometry Conjecture}

\paragraph{Concrete content.}
This candidate introduces a new proxy object called quantum chaotic geometry and uses it to connect quantum geometry with chaotic dynamics. The intended statement is that the geometric dimension of quantum states changes adaptively with chaotic evolution. The underlying intuition is that quantum effects and classical chaos might share a deeper geometric control variable.

\paragraph{Significance and novelty.}
The significance lies in the search for a unified geometric interface between quantum effects and nonlinear dynamics. The novelty comes from the proposed object quantum chaotic geometry rather than from another statement about spectral statistics in quantum chaos. The main limitation is that the object still remains at the level of name and analogy.

\paragraph{Potential impact and application significance.}
If a more operational version emerged, the candidate could affect quantum chaos, quantum control, geometric complexity analysis in quantum information, and the modelling of selected nonlinear quantum systems. A concrete scenario is a driven quantum system in which spectral statistics, entanglement growth, and orbit geometry all change simultaneously. At present it is difficult to determine whether these changes are manifestations of one mechanism. A rigorous quantum chaotic geometry could place those observations on one geometric variable. The help to human researchers would be primarily explanatory because the conjecture offers a route for turning parallel observations into one coherent geometric story.

\paragraph{Natural language statement.}
For every chaotic dynamical system object \(X\) with quantum effects, if \(X\) carries quantum chaotic geometry and if quantum geometric dimension, chaotic evolution, and nonlinear dynamics are observable, then the geometric dimension changes adaptively with the chaotic evolution.

\paragraph{Lean simplified statement.}
\begin{Verbatim}[fontsize=\small,breaklines=true]
forall X : Object,
  InClass "C019.object_class" X ->
  HasProxy "C019.proxy_object" X ->
  HasInvariant "C019.invariant.0" X ->
  HasInvariant "C019.invariant.1" X ->
  HasInvariant "C019.invariant.2" X ->
  AsymptoticRateRel "C019.relation" X
\end{Verbatim}

Foundationality is 4.0. Novelty is 3.0. Potential significance is 4.5.

\subsection{Case C020 Boundary Layers and Global Solutions in High Dimensional Fluid Dynamics}

\paragraph{Concrete content.}
This candidate concerns boundary layers and global solution behaviour in high dimensional fluid dynamics. The proposal is to use a high dimensional vortex and boundary proxy object to connect local boundary layer stability with global solution stability. The substantive claim is that local boundary mechanisms may be the dominant variables that determine global regularity, long term stability, or large scale structure.

\paragraph{Significance and novelty.}
The value of the candidate lies in the fact that one of the hardest questions in partial differential equations and fluid dynamics is how local boundary effects scale into global behaviour. The novelty lies in the introduction of a dedicated boundary proxy object, rather than a separate treatment of boundary layer estimates and global solution estimates. The main limitation is that the class of equations, boundary conditions, and proxy object remain open.

\paragraph{Potential impact and application significance.}
If a concrete version could be developed, the direction would affect boundary layer theory, criteria for global regularity, computational fluid dynamics, and stability analysis of high dimensional fluid systems. The significance comes from promoting local boundary mechanisms into principal explanatory objects for global behaviour. A concrete scenario arises in numerical simulation, where instability near a wall often appears before researchers know whether the effect will escalate into global instability, blow up risk, or long time structural change. A suitable boundary proxy object could function as the mediating quantity between local diagnostics and global prediction. The benefit to human researchers would not be only theoretical. It could also influence model reduction, grid resolution choices, and sensitivity analysis with respect to boundary conditions.

\paragraph{Natural language statement.}
For every high dimensional fluid system object \(X\) with boundary layers, if \(X\) carries a high dimensional vortex and boundary proxy object and if boundary layer stability, global solution stability, and vortex structure are observable, then boundary layer stability is structurally related to global solution behaviour.

\paragraph{Lean simplified statement.}
\begin{Verbatim}[fontsize=\small,breaklines=true]
forall X : Object,
  InClass "C020.object_class" X ->
  HasProxy "C020.proxy_object" X ->
  HasInvariant "C020.invariant.0" X ->
  HasInvariant "C020.invariant.1" X ->
  HasInvariant "C020.invariant.2" X ->
  StructureDecompositionRel "C020.relation" X
\end{Verbatim}

Foundationality is 4.5. Novelty is 3.5. Potential significance is 6.0.

The seventeen appendix cases show that a substantial amount of structural variation remains outside the three primary cases in the main body. Several cases such as C001, C002, C007, C008, and C011 already exhibit relatively clear proxy objects and mathematically meaningful tension, even though they are not selected as primary representatives. Other cases reveal the frequent limitations of cross domain analogy, namely unstable objects, unspecified mappings, and weak quantification. Taken together, the appendix cases extend the picture given by the three primary cases and provide a broader view of the boundary of the current pipeline.

\end{document}